# Element-Wise Attention Layers: an option for optimization

**Giovanni Araujo Bacochina**
**Rodrigo Clemente Thom de Souza**

**Federal University of Paraná**

## Abstract

The use of Attention Layers has become a trend since the popularization of the Transformer-based models, being the key element for many state-of-the-art models that have been developed through recent years. However, one of the biggest obstacles in implementing these architectures – as well as many others in Deep Learning Field – is the enormous amount of optimizing parameters they possess, which make its use conditioned on the availability of robust hardware. In this paper, it's proposed a new method of attention mechanism that adapts the Dot-Product Attention, which uses matrices multiplications, to become element-wise through the use of arrays multiplications. To test the effectiveness of such approach, two models – one with a VGG-like architecture and one with the proposed method - have been trained in a classification task using Fashion MNIST and CIFAR10 datasets. Each model has been trained for 10 epochs in a single Tesla T4 GPU from Google Colaboratory. The results show that this mechanism allows for an accuracy of 92% of the VGG-like counterpart in Fashion MNIST dataset, while reducing the number of parameters in 97%. For CIFAR10, the accuracy is still equivalent to 60% of the VGG-like counterpart while using 50% less parameters.

## 1. Introduction

Neural Machine Translation is a technique for translation that was proposed by Kalchbrenner and Blunsom in 2013[1]. This technique basically consisted of training a model to extract features from words in a reference language and relate those features with their respective translation in the target language. Its main innovation was using a single model to perform the mentioned task, instead of many sub-components.

For this task, the main architecture used was an encoder-decoder composed of Recurrent Neural Networks. The encoder would receive the reference language, extract its features, decomposing them into a fixed-length vector. Then, the decoder would receive this vector and recompose it into a translated output. However, one problem with this approach is that the fixed-length vector might be too small for the features that can be extracted from the input, which caused the model performance to be greatly reduced when dealing with long sentences.

To avoid the loss of performance caused by this bottleneck, an "annotation" mechanism has been proposed in 2014 by Bahdanau *et al*[2], by which the Encoder would become a bidirectional RNN and would assign a value to each input sequence. The weighted sum of all those annotations would be assigned to a context vector, which composed the fixed-length vector to be passed as the encoder output. The weight of such sum would be defined by a feedforward neural network, which made it subject to optimization through gradient descent methods. Such approach provided a model capable of generating outputs with a consistent quality while also being resilient to the input sentence length. This was the creation of what would be called Attention Mechanism.

The effectiveness of this approach was soon embraced by the popular Seq2Seq model, proposed by Mikolov.

However, despite such effectiveness, the initial problem wasn't solved in fact, but only mitigated.
This motivated the creation of the Transformer by Vaswani *et al.* in 2017[3], a model which discarded the use of Recurrent Neural Networks and was based entirely on matrix multiplications, which resulted in fewer parameters, easier parallelization and faster training. All this while achieving a new state-of-the-art in translation quality.

The fundamental of Transformer is summarized by its paper title: "Attention is All You Need". In fact, by using the Dot-Product Attention layer and Self-Attention, mechanisms fairly simple, the model could produce outputs with a Bleu Score of 41.8[3] when translating English to French after being trained on news-test-2014, while Badanau *et al*.'s RNNsearch-50 performance stagnated before surpassing the 30 mark[2].

In fact, Transformer was good not only for translation. Using exclusively the Encoder part in a bidirectional version, Devlin *et al*. developed BERT, a powerful model for text classification[4]. With the Decoder, Radford *et al*. created the Generative Pretrained Transformer, or GPT, a model focused on generating texts and which gave birth to the GPT-2, GPT-3 and the newest ChatGPT, which, despite its problem with misinformation, can generate impressive human-like, long texts[5][6][7].

However, new approaches outside Natural Language area have been developed for multiple tasks using the ideas proposed with Transformer, such as object identification[9], movements classification[10], data generation[8][11] and even in Reinforcement Learning to play games[12].

These projects show how powerful the attention mechanism is. However, a notable disadvantage of this technique is that, due to fact that it resorts to matrix multiplications, it tends to generate too many optimizing parameters, especially when dealing with inputs that has big amounts of data, like images.

For this motive, Goodfellow *et al*. adapted Transformer's MultiHead Attention to be a 2D Convolution with kernel size 1x1, instead of a Fully Connected layer[8]. This approach, though keen, decreases severely the training and inference speed.

In this work, it's proposed another alternative, one that tries to be more similar to the way the original MultiHead Attention works, but instead of relying on matrix multiplications, it relies on arrays multiplications, which are element-wise. This way, it's possible to assign weights to each value in the input array in relation to other values located in both the same row and the same column. Thus, this method is called Attention-Wise.

In order to test the effectiveness of this method, an attention model has been trained on MNIST and CIFAR10 using different configurations which will be discussed in this work, being compared with a VGG-like model which has 7 layers. The outputs have been visualized in order to check how the mechanism was behaving and also to test possible improvements.

At the Conclusion Section, it's discussed some possible disadvantages of this method, as well as some considerations regarding the relation between hardware leverage and optimization.

At the end of this paper, there's an Appendix featuring additional experiments that aim to test some of the discussed issues that this technique might include.

## 2. Methodology

In this section, it's discussed more details on the Control Model architecture and its number of parameters. Then, the different configurations that were tested for the

Attention Model, as well as their respective number of parameters. Lastly, it's stablished the hyperparameters that were used for the training process.

## 2.1 The Control Model

The Control Model architecture is composed of 6 Convolution layers followed by a Fully Connected layer, resembling a VGG model. The motive to why this architecture has been chosen is due to it being elementary and intuitive, being the easiest one to learn and comprehend when studying classification tasks using images and feature extraction from images in general.

The first convolution layer has kernels 3x3 and padding 1, generating 100 feature maps. The second layer also has 3x3 kernels and padding, but keeps the number of feature maps. The third layer has kernels 2x2, with stride 2 and no padding, thereby downsampling the input. The fourth follows the same pattern as the first one, generating 200 feature maps. The fifth, the same pattern as the second. Finally, the sixth convolution downsamples the input again. No bias has been used in the downsampling convolutions. The final layer, then, is a Fully Connected layer, which generates 10 outputs to be passed through a Logsoftmax function.
Every conv layer has been followed by a Rectified Linear Unit as activation function.

Considering that all code has been deployed using Pytorch, which by default(argument groups=1) makes all inputs being convolved to all outputs, the number of parameters for each convolution can be calculated as:

$$\sum parameters = input\_channels \cdot output\_channels \cdot (kernel_i \cdot kernel_j)$$

*( 1 )*

Where i and j represents the kernel rows and columns, respectively.

Meanwhile, considering that a Fully Connected Layer consists of a matrix multiplication, the number of parameters will be given by:

$$\sum parameters = n\_input\_features \cdot n\_output\_features$$

*( 2 )*

Therefore, we can calculate that our Control Model for classifying the Fashion MNIST[15] dataset will have **929,510** optimizing parameters. For CIFAR10[16], after making the proper adaptations, this number raises to **961,310** parameters.

Those numbers are far from the more than 130 million possessed by VGG variants, but it's still a reasonable amount for the selected datasets and for the desired purposes.

## 2.2 The Attention Model

The Attention Model is composed of attention-wise layers, each layer having multiple heads. The model was tested with 2 and 4 attention layers, with 8 and with 16 attention heads.

Each Attention Layer includes multiple attention heads, which will generate a single output each. This output is then concatenated in the channels dimension, providing an output with channels = channels · n_heads. In order to filter the best suitable outputs while also recomposing the number of input channels, this output is passed through a convolution layer. This convolution output is then summed to the attention layer input. After this, there might be a batch normalization layer before passing the tensor through a ReLU function and returning its output.

In session 3 will be demonstrated that this batch normalization layer can be removed to actually improve the model performance without compromising the backpropagation process.

The Attention-Wise mechanism was developed based on the Scaled Dot-Product Attention used in Transformer[3], with some substantial changes due to the type of data used in this situation. Since the input and the output are not in the form of vectors, but rather in 3 dimensional arrays, there are no queries, keys nor values vectors. Those have been replaced by 2 arrays of weights, one representing the arrays for the rows in the input data, thus being referred to as **x weights** array, and another for the columns, referred to as **y weights** array. Each weight array has the same number of dimensions as the input data. Therefore, for Fashion MNIST, all weights have 1 channel, 28 pixels as width and 28 as height. For CIFAR10, 3x32x32.

Therefore, considering an input of height 3, width 3 and a single channel, its x weights and y weights will be given by:

$$\begin{matrix} Wx_{11} & Wx_{12} & Wx_{13} \\ Wx_{21} & Wx_{22} & Wx_{23} \\ Wx_{31} & Wx_{32} & Wx_{33} \end{matrix} \qquad \begin{matrix} Wy_{11} & Wy_{12} & Wy_{13} \\ Wy_{21} & Wy_{22} & Wy_{23} \\ Wy_{31} & Wy_{32} & Wy_{33} \end{matrix}$$

Where $Wx_{ij}$ represents the x weight at the row **i** and column **j**, the same being applied to $Wy_{ij}$.

Each weight is initialized from a gaussian distribution with a standard deviation of 0.02. The element-wise product of a weight array by the input array is passed to a softmax function, providing a value within range [0,1], thus, providing the degree of relevance of each value in the input.

For the x weights, the softmax function is applied over each row of the product array, while for the x weights, over each column. This operation make it possible to assign a degree of relevance of each value in relation to all other values within the same row(for x weights product) or within the same column(for y weights product):

$$\begin{pmatrix} I_{11} & I_{12} & I_{13} \\ I_{21} & I_{22} & I_{23} \\ I_{31} & I_{32} & I_{33} \end{pmatrix} \cdot \begin{pmatrix} Wx_{11} & Wx_{12} & Wx_{13} \\ Wx_{21} & Wx_{22} & Wx_{23} \\ Wx_{31} & Wx_{32} & Wx_{33} \end{pmatrix} = \begin{pmatrix} Ox_{11} & Ox_{12} & Ox_{13} \\ Ox_{21} & Ox_{22} & Ox_{23} \\ Ox_{31} & Ox_{32} & Ox_{33} \end{pmatrix}$$

(3)

Where the element $Ox_{ij}$ is given by:

$$Ox_{ij} = I_{ij} * Wx_{ij}$$

(4)

If softmax is applied through the **i** row, the output will be given by:

$$Ox = \frac{e^{Oxi}}{\sum_j e^{Oxj}}$$

(5)

For the y axis, this output will be slightly different, with the softmax being applied through the **j** column:

$$Oy = \frac{e^{Oxj}}{\sum_i e^{Oxi}}$$

(6)

This operation will provide one output array for the x axis, consisting of relevance degree for each value in each row, and an output array for the y axis, consisting of the same factor, but for each column.

Since this process is done considering the relevance degree of each element in relation to the neighboring values, it could be somehow compared to the convolution process, but discarding kernels and being a much faster process.

After the softmax outputs for both rows and columns have been obtained, they are multiplied together in order to properly obtain a relevance array, which considers the relevance of each feature in relation to its neighbors both in the same row and the same column.

$$Out = Ox * Oy$$

(7)

Note that the most relevant feature will receive a higher relative relevance value for both Ox and Oy, while the less relevant will receive a lower value. Intermediary values might receive a higher value for Ox but lower for Oy and vice-versa. This allows for a compensation should an important feature receive a lower relevance on its current row or in its current column.

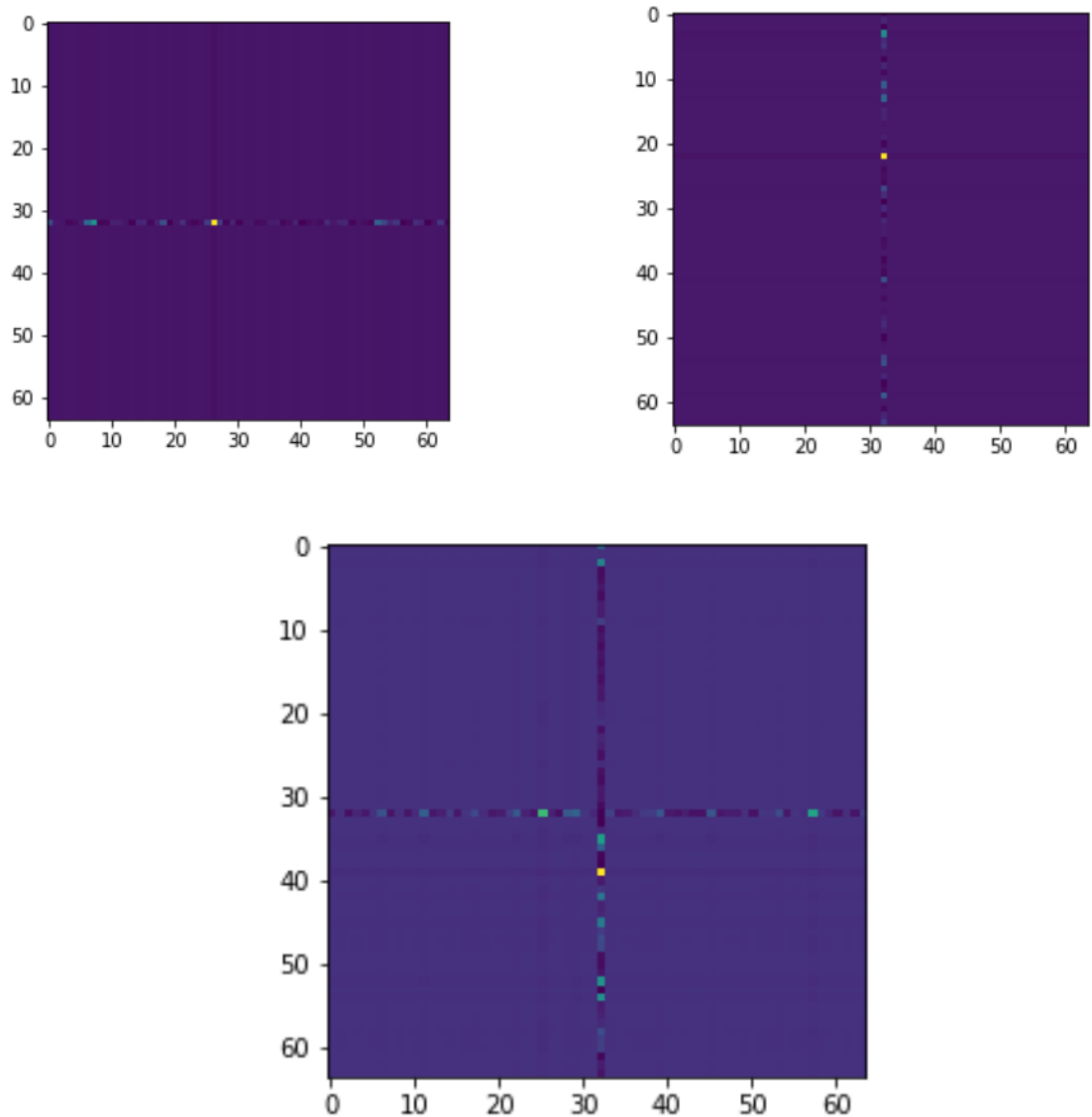

*Figure 1: illustrating how the Attention-Wise weights work. Each image represents a weight array. The higher the relevancy of a certain pixel in relation to its axis, the higher its weight value, thus, the lighter its color.*

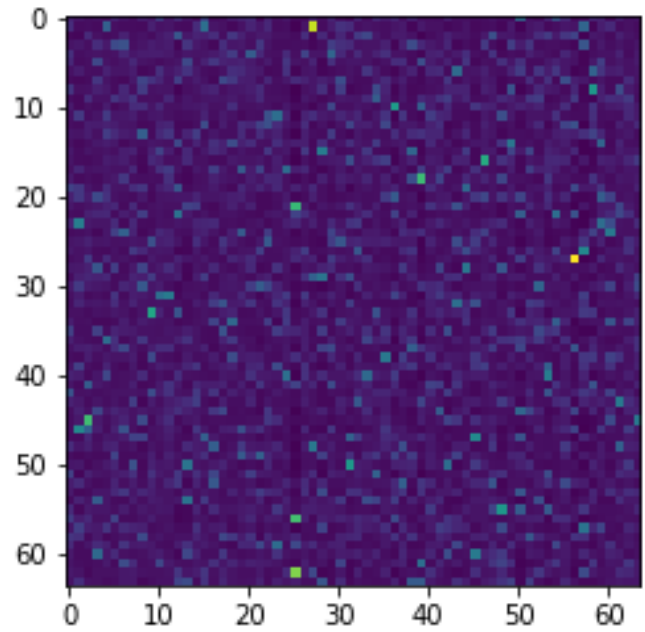

*Figure 2: A possible weight array after applying the product of X weights and Y weights. Note that values close to 1(yellow color) are relatively scarce, which can help avoid problems related to overfitting and "memorization".*

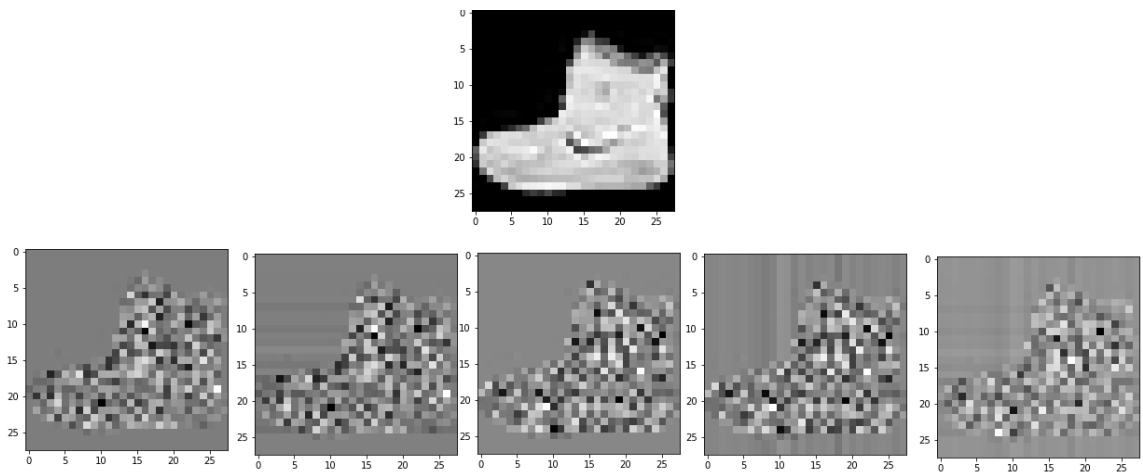

*Figure 3: The input image(up) and the outputs provided by a trained X weights; the same output after being passed through a softmax; by Y weights; after softmax; the final output. Note the stripes that are added by the softmax.*

In the end, the array of X and Y weights is passed through a batchnorm layer, which serves as a replacement for the scaling factor $1/\sqrt{dk}$ used in Scaled Dot-Product Attention, only then being applied to the input data through array multiplication.

Note that all multiplications are done element-wise as any array multiplication.

A remark that must be done is that there's only one x weight array and a single y weights array, being those defined when the model is initialized. This grants that those weights will be generalist and avoid the possibility of overfitting.

Since the Attention-Wise mechanism was tested using multiple heads, the outputs of each head is concatenated in the channel dimension and then passed through a convolution layer, which provides an output with the desired number of channels. In the experiments, this number is always the same as the original images.

The convolution attention is also summed to a skip connection, passed through a batch normalization layer in some cases, and, finally, a ReLU function is applied, providing the final output. In the extra experiments, it was noted that removing the skip connection increases instability during backpropagation while not providing a meaningful increment in performance.

After the attention layers, there's also a fully connected layer to provide the classification output.

The number of parameters of an Attention-Wise layer can then be calculated as:

$$\sum parameters = (input\_channels \cdot input\_width \cdot input\_height \cdot 2 \cdot n\_heads) + Conv\_parameters$$

*( 8 )*

Where Conv_parameters has been defined in (1).

The different configurations for the Attention Model that have been tested are listed below with its respective number of parameters:

**Fashion MNIST dataset:**

- ➔ A) 8 Heads + 2 Attention Layers with BatchNorm + FCC layer = 33,104
- ➔ B) 8 Heads + 2 Attention Layers without BatchNorm + FCC layer = 33,084
- ➔ C) 16 Heads + 2 Attention Layers with BatchNorm + FCC layer = 58,208
- ➔ D) 16 Heads + 2 Attention Layers without BatchNorm + FCC layer = 58,172
- ➔ E) 8 Heads + 4 Attention Layers with BatchNorm + FCC layer = 58,342
- ➔ F) 8 Heads + 4 Attention Layers without BatchNorm + FCC layer = 58,318
- ➔ G) 16 Heads + 4 Attention Layers with BatchNorm + FCC layer = 108,534
- ➔ H) 16 Heads + 4 Attention Layers without BatchNorm + FCC layer = 108,494

**CIFAR10 dataset:**

- ➔ 8 Heads + 4 Attention Layers without BatchNorm + FCC layer = 506,560

The remarked BatchNorm layer is the one that follows the convolution, as mentioned above.

### 2.3 The Training Configuration

Each model had its weights initialized using a gaussian distribution with a standard deviation of 0.02. The biases were initialized with value 0.

All models were trained for 10 epochs. For Fashion MNIST, the batch size was 4096. For CIFAR10, 1024.

For optimization, Adam optimizer was chosen with default hyperparameters – $\beta_1$ = 0.9, $\beta_2$ = 0.999, learning rate = 1e-3.

## 3. Experiments

Initially, the Control Model and the Attention Model in the configuration A have been trained together. The results obtained for the former have been saved and used as reference.

The metrics that have been used were the Loss value using Categorical Cross-Entropy on training data and Accuracy on test data. For better comprehension on each model's behavior, the gradients average of the first layer – the first convolution layer for Control Model, the x weights array in the first head for Attention Model – have also been measured.

The experiments have been performed in a single GPU Tesla T4 provided by Google Colaboratory. The training process took around 35 minutes for Attention Model A and B, about 70 minutes for C and D, 1 and a half hour for E and F, and 3 hours for G and H.

In Table 1 are the metrics obtained after 10 epochs.

|  | Loss | Accuracy(%) | Last Gradients Average |
|---|---|---|---|
| Control | 4.508300 | 88.290000 | 0.002000 |
| Attention A | 34.539000 | nan | 0.000000 |
| Attention B | 7.591300 | 81.460000 | 0.000029 |
| Attention C | 34.538800 | nan | 0.000000 |
| Attention D | 7.485600 | 81.660000 | 0.000001 |
| Attention E | 9.852200 | 77.090000 | 0.000032 |
| Attention F | 7.480600 | 81.330000 | 0.000004 |
| Attention G | 10.034500 | 76.370000 | 0.000055 |
| Attention H | 7.747500 | 80.790000 | -0.000006 |

*Table 1: The observed metrics with each model configuration after 10 epochs. NaN means that the optimizer failed to converge and the model collapsed.*

Note that the Attention Models A and C collapsed, not being able to learn any relevant feature. Their gradients values oscillated between 1e-12 and 9e-11 and the accuracy got stuck on 9.99% for the entire process. It's also interesting to remark that the only difference between models A|B and C|D is the fact that B and D didn't have batch normalization layers after the convolution layer.

This raised the hypothesis that the batch normalization layer could be sabotaging the attention mechanism performance rather than helping.

In order to test this hypothesis, it has been plotted the output of: 1) each attention layer as it is, before any operation; 2) after the convolution layer; 3) after the addition of the residual connection, right after the convolution and before the batch normalization; 4) right after the batch normalization layer; 5) the final attention layer output, after ReLU activation.

The results are shown in the figures 4, 5, 6, 7 and 8.

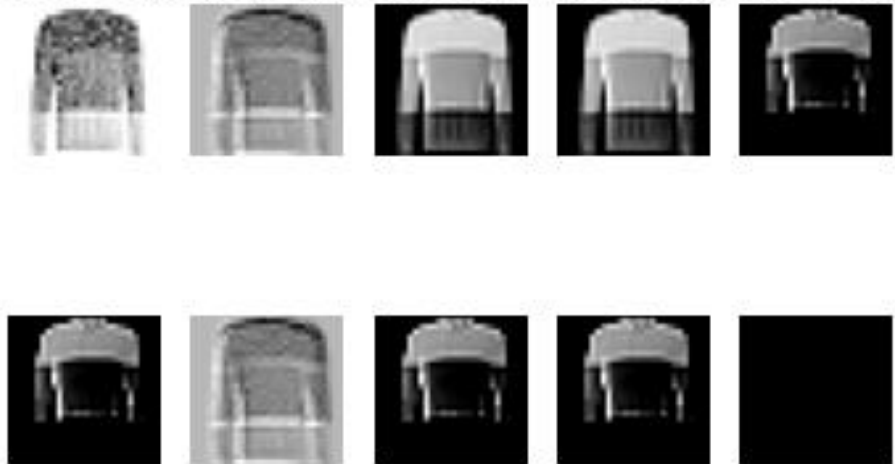

*Figure 4: The outputs provided by each operation by the model A. "Before Norm" is right after being added a skip connection, after the convolution layer and before the batch normalization layer.*

Figure 4 allows one fundamental conclusion: that the attention mechanism is in fact being optimized and the weights are having their values correctly changed. However, due to this model performance, it's still not possible to conclude if this mechanism is being efficient. In fact, the model collapse shows that the mechanism might actually be just copying and pasting a determined input image.

It's also noticeable that the ReLU activation function, in this case, tends to lead to information loss.

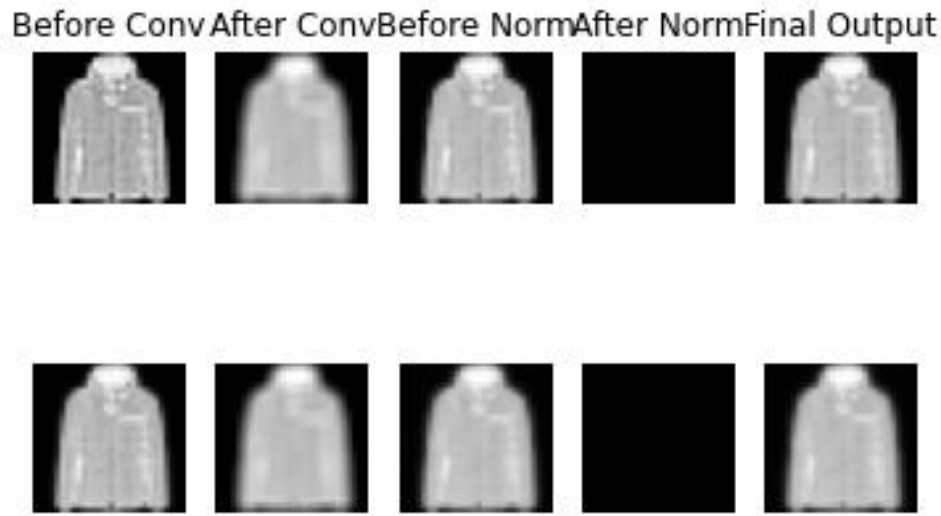

*Figure 5: The model B outputs. Note that, since model B has no batch normalization layer, the "After Norm" column corresponds to a zeros tensor.*

In Figure 5 one relevant trait is how the attention layer reproduces the object reliably, while the convolution tends to add blur, an artifact that is mitigated by the skip connection.

It's also interesting to note that, while this could also raise another warning around the possibility of overfitting or simple "copy-paste"(or redundancy), the fact that this model achieved a loss of 7.59 and an accuracy of 81.46% allows us to consider that the mechanism might actually be working in a similar way of a convolution.

Of course, it's still necessary to admit the scenario where such performance might not come from the attention layer, but rather from the feedforward layer.

We can also note here that the ReLU did no harm to the output at all.

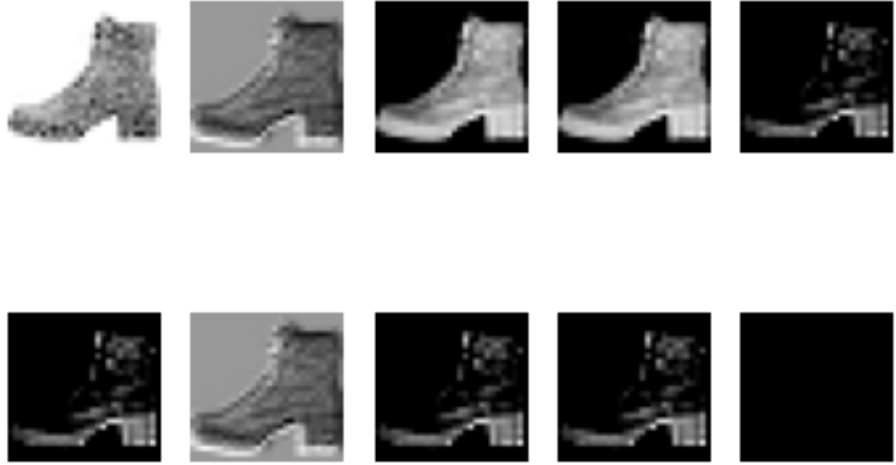

*Figure 6: Model C outputs. Note how some black and many shades of gray dots are distributed throughout the boot in the first image, demonstrating some pixels that were classified as having low relevancy. In this case, the convolution managed to mitigate such flaws.*

So far, the most notable change doesn't happen in the batch normalization output, but actually in the output that comes after the ReLU activation function.

When printing the output values, it was noted that, the attention heads tend to produce outputs which have pixel values within range [0.4, 0.5], where 0.5 usually represents pixels that compose the background – which, in the input image, have value 0.

When such output is passed through the convolution layer, this range is changed to [0.53, 0.70], where the background values receive the lowest value of this range(in this sample, 0.53112763). The convolution, in this matter, seems to act in a similar way than a pseudo-label assigner model[13], providing a better segregation between pixels that are located in different parts of the input(object shape, background pixels, shadows, pockets, neckband, details, etc) and assigning them values according to their "category", thus minimizing the image entropy in a continuous way. However, this is just a hypothesis and it hasn't been tested in this paper.

After the convolution provides a better segregation, the batch normalization would put them close together again, but generating many negative values in this process. As a result, the ReLU activation causes a great loss of data. One possible way to avoid this problem would be removing the ReLU activation, but if the pseudo-label hypothesis is right, it might be possible that we wouldn't be able to discard background pixels and focus entirely on objects.

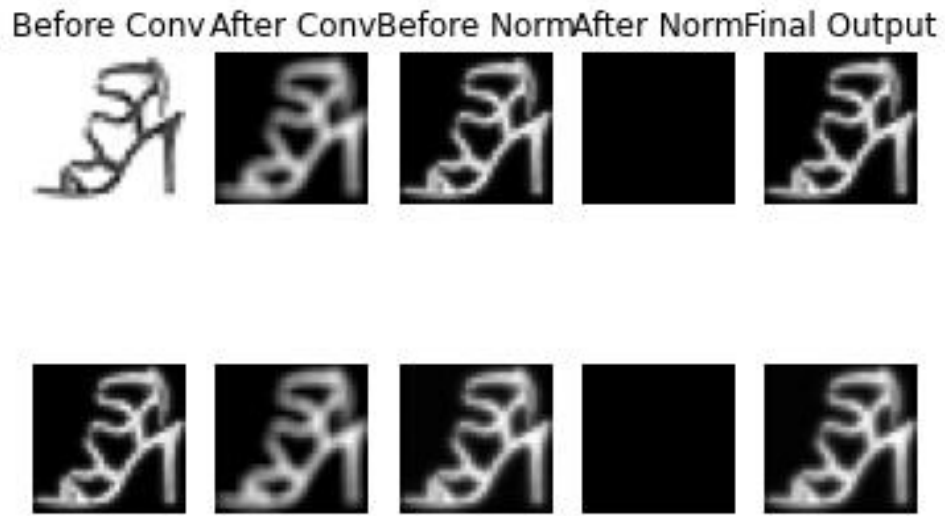

*Figure 8: The Model D outputs*

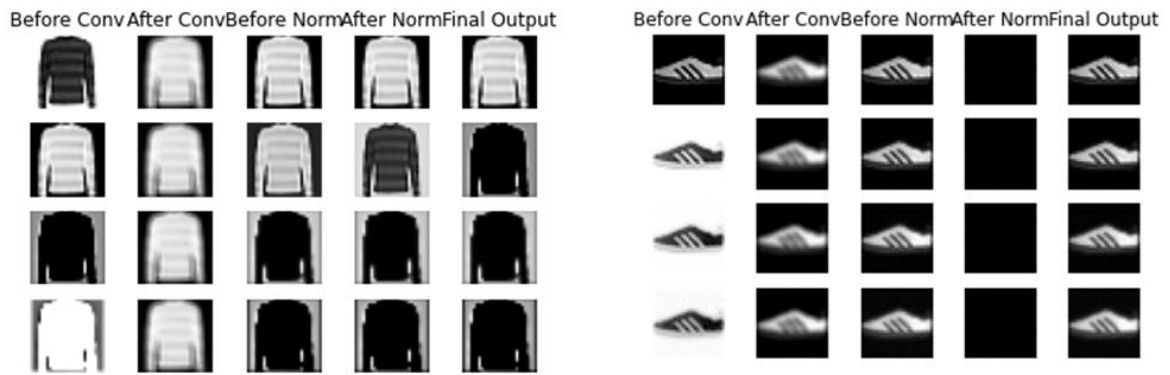

*Figure 7: Models E(left) and F(right) outputs.*

Finally, the loss, accuracy and gradients average were plotted for the Control Model and the attention A and B and the results can be seen in Figure 9. The X axis represents the number of iterations except for the accuracy plot, where it actually represents the number of epochs:

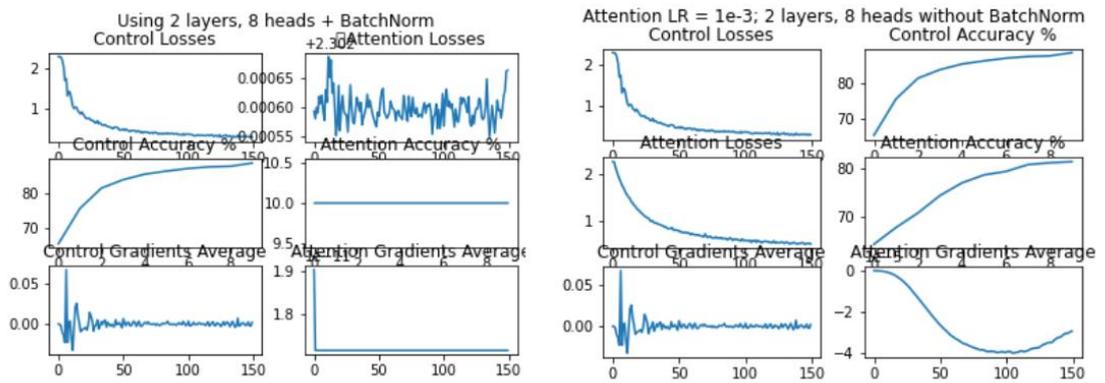

*Figure 9: At the left, the graph abnormal behavior shows that the attention model failed to be optimized. At the right, the attention model manages to successfully learn how to correctly classify the data. The control gradients act normally, having bigger values.*

The metrics behavior was similar for the others configurations of the Attention Model.

Finally, the x weights and the y weights of the first attention head of model G were printed to check whether they were simply trying to mimic the input image or simply output it without any modification. Most values in those arrays were within range [-0.05, 0.05], with the y weights having a smaller standard deviation, being common to see values around -1e-3 and -1e-4.

In the end, to measure the attention-wise effectiveness in a more complex dataset, an Attention Model with 4 layers and 8 heads was trained on CIFAR10 dataset without batch normalization layer after the convolution.

The results are plotted in Figure 10.

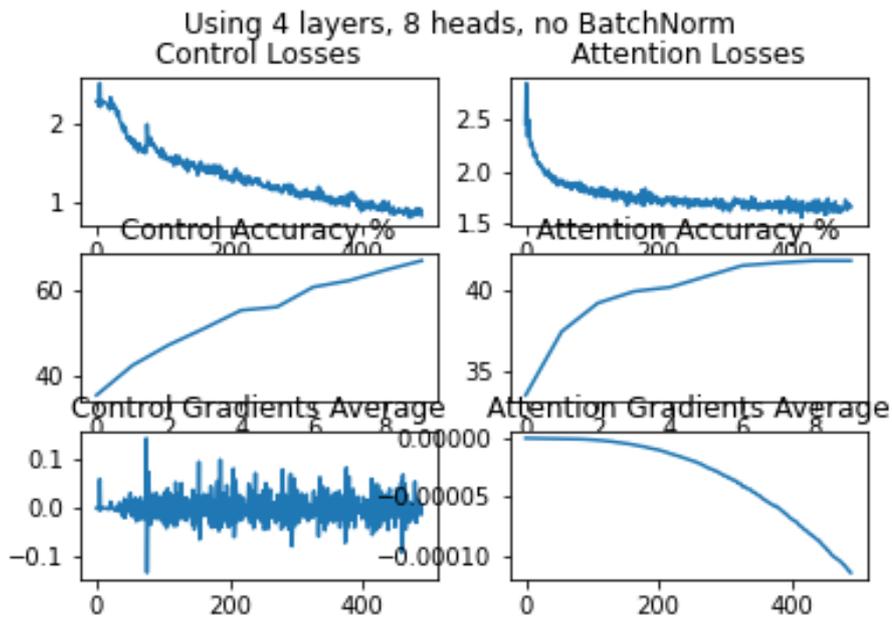

*Figure 10: Though the loss and accuracy had reached their plateau, the gradients were still moving away from zero.*

After 10 epochs, the Control Model achieved a loss of 42.56 and an accuracy of 67.12%. Its last gradients average was 0.0062. Meanwhile, the Attention Model achieved a loss of 80.89, accuracy of 41.81% and its gradients average was 0.0001.

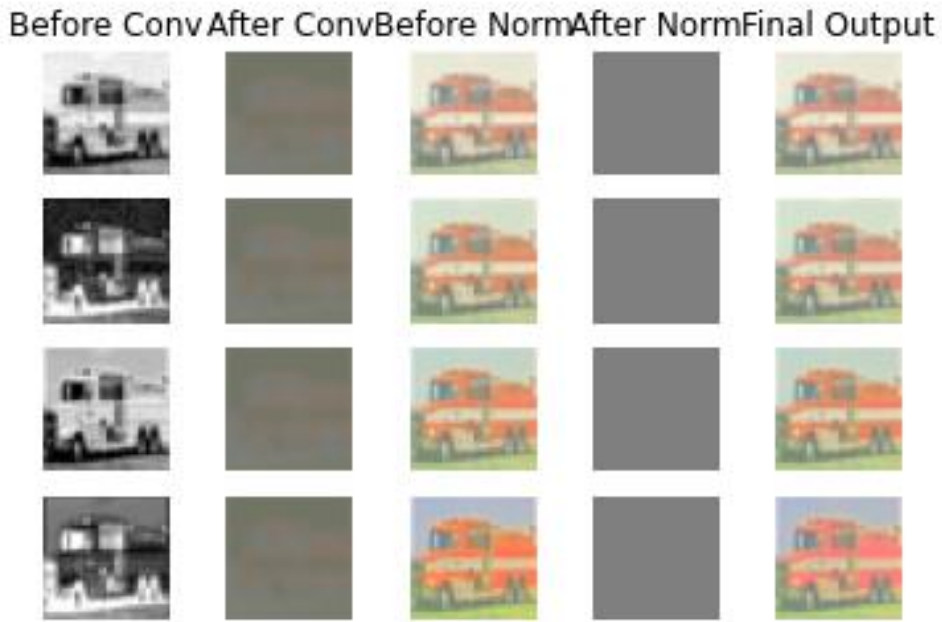

*Figure 11: The outputs of the attention model for each layer. Remember that there's no output from the batch normalization layer, thus, it was used a single zeros tensor.*

In this case, the convolution tends to generate an output with values within range [0.4, 0.5], which makes it barely visible(Figure 11). The performance metrics, however, indicate extreme overfitting. Figure 10 also shows that the Attention Model reaches its plateau of performance way before the control model, suggesting that way more parameters are necessary to achieve a performance comparable to the Control. Thus, the main advantage of this technique, which is optimization, might be compromised.

## 4. Conclusion

In this paper, it's presented the Attention-Wise layer, an attention mechanism which relies on arrays multiplications in order to assign weights to input arrays while also avoiding dealing with too many parameters and bypassing the increasing training time issue that is characteristic to convolution layers with too many channel outputs.

It was shown that this mechanism can achieve a similar performance that of a model with a VGG-like architecture with way less parameters and faster training, avoiding or at least mitigating the need for hardware leverage for some tasks. Even though Richard Sutton states that hardware leverage and higher performance are intrinsically connected[14], it's always important to look for techniques that might make Deep Learning less computationally expensive, so such technology can be better disseminated through society, improving the relation cost-benefit for applying DL at different sectors.

And, of course, the optimization allows for those who are able to enjoy more powerful hardware to be able to do even more with their resources.

It's also remarkable that the Attention-Wise technique, due to relying on 2 applications of the softmax function, generates small gradients which are even smaller if the x and y weights are initialized with extremely low values, as can be seen in the plots. This can make performance gain through training too slow. However, increasing the learning rate to 0.1 caused the collapse of the model, even without batch normalization, possibly because of the convolution and the feedforward layers. A possible bypass to this issue would be optimizing the attention layers separately from the convolution and feedforward layers.

Another remark that must be done is that, due to the fact that the Attention-Wise technique was developed inspired by MultiHead Attention, its architecture includes the use of multiple heads. However, the results show that using multiple heads is of little relevance to the model performance, as can be perceived when comparing models B to D and F to H in Table 1.

In order to confirm and check ways to avoid the issues stated above, extra experiments were made, which are available at the Appendix.

Lastly, perhaps the main issue that can be observed in this work is that, as the models B and F demonstrated, using more Attention Layers does not necessarily provides a change in the performance. Such issue can be a limiting factor in the Attention-Wise mechanism. Solving this problem can be the key to make such technique a viable alternative to the default methods used in Deep Learning such as convolutions and matrix multiplication.

To conclude this study, some ideas could rise up in order to make good use of the attention-wise mechanism. A possible upgrade would be using bias, something that wasn't applied in this paper. An interesting approach would be comparing the performance of such method on upsampled and downsampled inputs. Another promising project would be testing attention-wise layers for object detection and image segmentation. Generative models could also benefit greatly from attention-wise layers, though it might be troublesome for GANs due to instability issues – for those who consider testing this option, prefer using the attention layer at the end of the generator, as demonstrated by SAGAN[8].

# Appendix: Extra Experiments

Here are presented extra experiments in order to avoid some of the encountered issues of the attention model while testing the possibilities discussed at the end of section 3. Another experiment has also been performed in order to check the model performance when the skip connection is removed.

The configurations that were used are:

**Fashion MNIST) 1 Head + 2 Attention Layers with BatchNorm + FCC layer = 11,132**

**CIFAR10) 8 Heads + 4 Attention Layers without BatchNorm + FCC layer = 506,560 → Optimizer A for Attention weights, Optimizer B for Conv and FCC layer.**

The first experiment intended to test how the Attention Model would perform when using just a single head and whether the number of heads could be a hyperparameter or not.

The third experiment, for the attention model without skip connections, was performed in the Fashion MNIST and in CIFAR10 using the configurations from model B(2 Attention layers, 8 Heads, no batch normalization).

Figure 12 shows the results for the single head Attention Model.

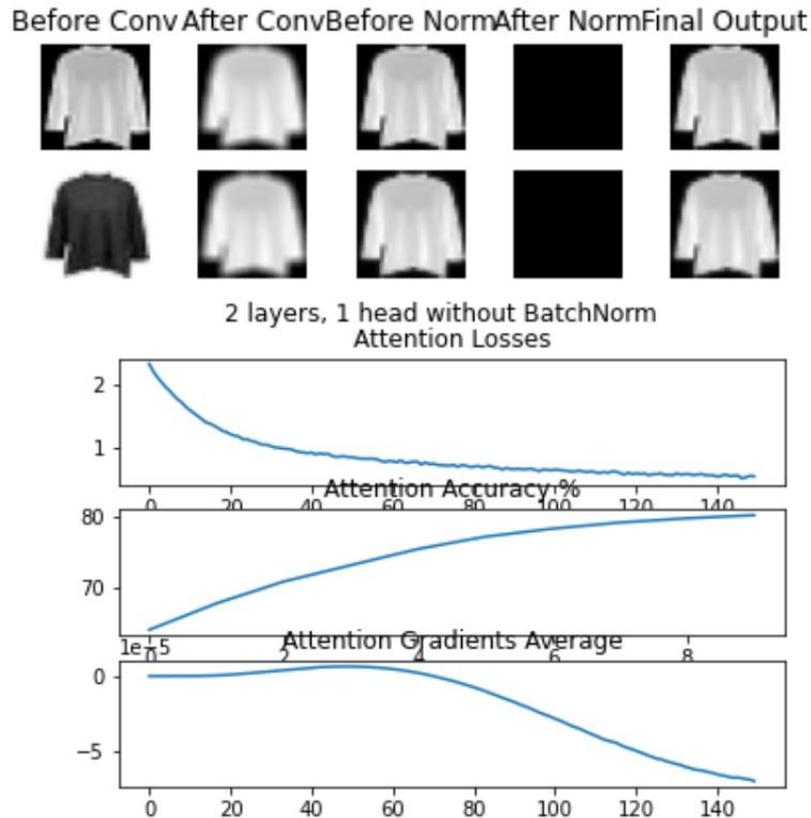

*Figure 12: Note how using a single head increased the slope in all graphs, but the results were pretty much the same.*

The final loss of this model was 8.3712, the accuracy was 80.09% and the gradients average, 6.9391e-5. Comparing with model D in Table 1 – loss 7.4856 and accuracy 81.66% - it's possible to conclude that using more than one attention head is a trivial matter, not providing significant increase in performance.

The second experiment was intended to check if optimizing the attention heads independently from the convolution and the feedforward layer would work as a bypass to the training time issue, since the attention-wise mechanism tends to generate extremely low gradients and a higher learning rate can lead to collapse of the convolution and feedforward weights. With such intention, it was created 2 lists of parameters from the attention model, one with all attention-wise layers and one with all the remaining parameters. Pytorch makes it able to pass each list to an independent optimizer.

The optimizer for the convolution and feedforward layers is the Adam optimizer with the same hyperparameters as used in the experiments described before. For the attention weights, however, it was used a learning rate of 0.1 instead of 0.001.

The results are as demonstrated in Figure 13.

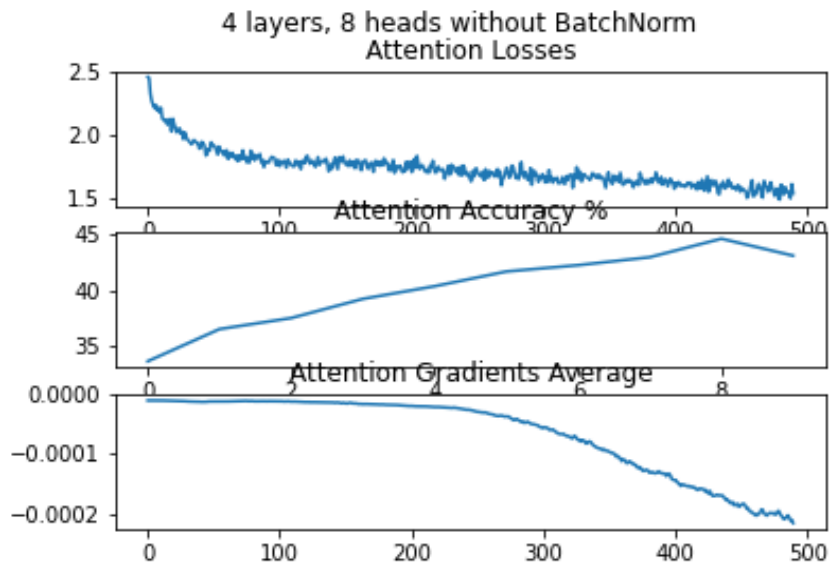

*Figure 13*

After 10 epochs, the model achieved a loss of 76.4777 and an accuracy of 43.10%. It reached its peak of accuracy at 9th epoch with an accuracy of 44.63%. Its last gradients average was -0.0002. Therefore, increasing the learning rate didn't provide relevant gains to the model performance, but made it reach its peak earlier. It can be assumed, then, that using segregated optimization might overcome the problem of low gradients and make it possible for the attention layers to reach their peak of performance at the same rate as the convolution and feedforward layers.

However, the outputs generated through this process are curiously different from the ones generated in the previous model. Figure 14 shows that the attention layer might be efficiently selecting the focused object, while the skip connection after the convolution appear to highlight some small details that either weren't captured by the attention or simply were faded by the convolution.

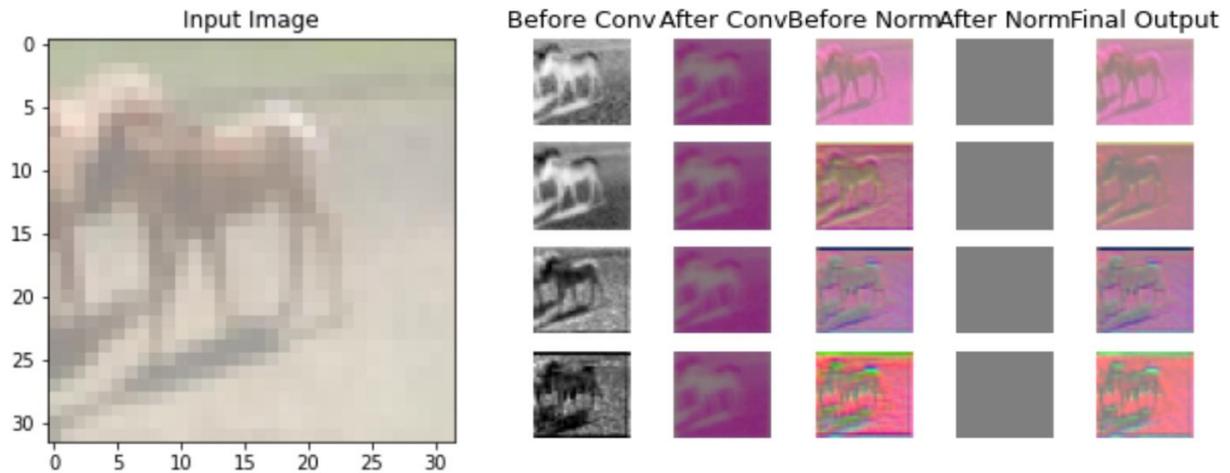

*Figure 14: There seems to be 2 dogs at the input image, yet only 1 has been highlighted by the model. Note how the model mainly assigns the color purple and red to the background, while highlighting some details in green and blue.*

Finally, the third experiment, which discarded the use of skip connections in an attention-wise model with 2 layers and 8 heads and its results are shown in Figure 15.

The first observation that could be made during this experiment is that removing the residual connection makes the model too unstable, even without Batch Normalization. It was necessary to use segregated optimization, manually selecting the best hyperparameters for optimizing the attention layers.

For MNIST, it was necessary to use a learning rate of 0.1 for the attention optimizer. Note that the Adam optimizer that optimized the convolution and the fully connected layers was kept with the default hyperparameters.

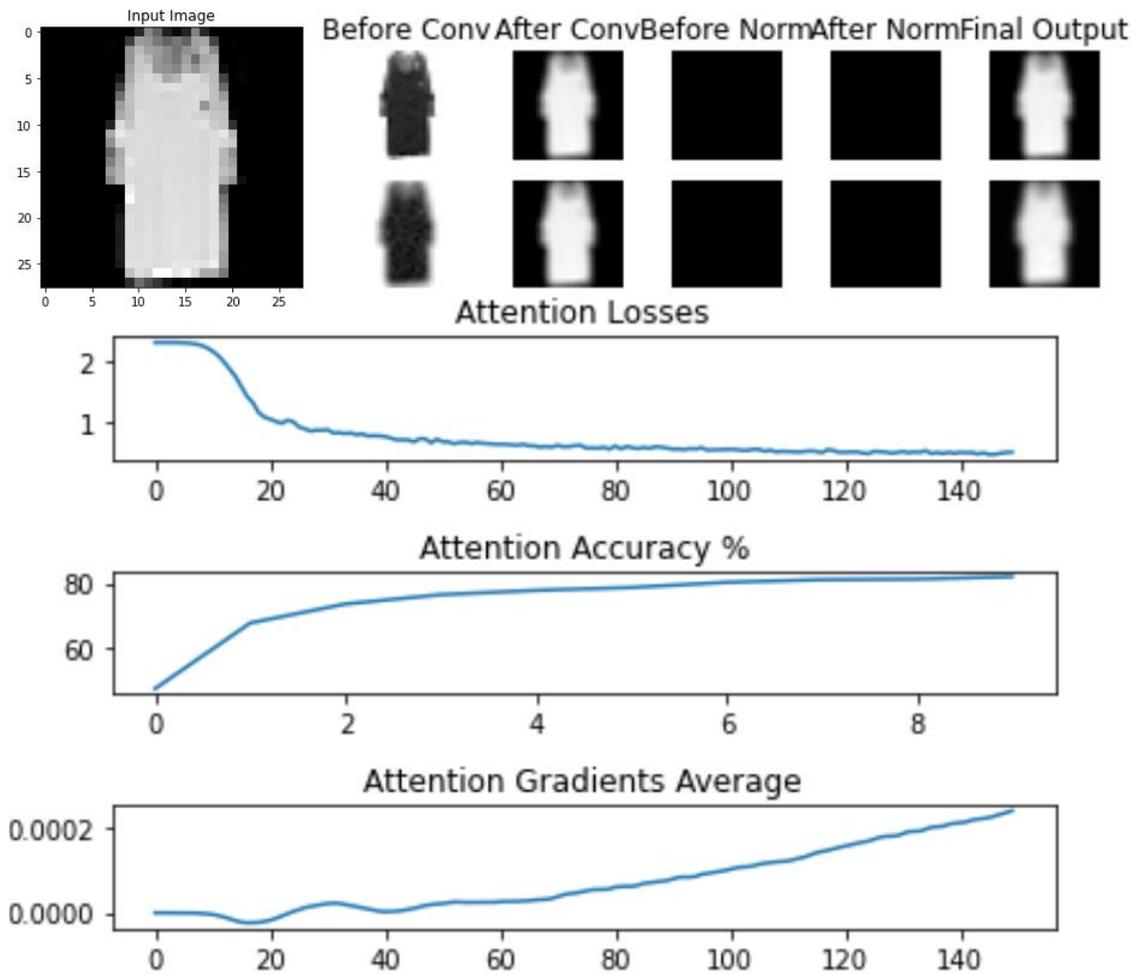

*Figure 15: Discarding the residual connections make the blur added by the convolution layer to be passed through all the following operations.*

Note that the cause to such instability is because of how the gradients behave: they begin near zero, and only as the weights start becoming considerable values the gradients begin to grow. This might indicate that the gaussian initialization may be inappropriate for this method.

However, it's interesting to note that even without skip connections the model managed to adapt and reach a loss value of 7.4980 and an accuracy of 82.13%. Its last gradients had an average of 0.0002 and the attention output show a good quality even in relation to the original input one. When the X weights of the first head in the first layer were checked, their mean was of 0.0007 and had a standard-deviation of 0.0208. For the Y weights, those values were 0.0004 and 0.0198 respectively.

For CIFAR10 dataset, initializing all layers with a gaussian distribution with mean 0 and standard deviation of 0.02 caused vanishing gradients and the optimizer failed to converge. However, initializing the convolution and feedforward layers with a normal distribution resulted in exploding gradients in the attention layer, with the loss value consistently staying above the mark of hundreds of millions, accuracy reaching its peak at 10.05% at the 3$^{rd}$ epoch and the gradients average consistently around the value of -230,000. Using smaller learning rates was ineffective at stabilizing the model. Initializing the attention layers with higher values also failed at the task. Thus, it wasn't possible to properly stabilize the model without skip connections for CIFAR10.